\documentclass[10pt,twocolumn,letterpaper]{article}

\usepackage{iccv}
\usepackage{times}
\usepackage{epsfig}
\usepackage{graphicx}
\usepackage{amsmath}
\usepackage{amssymb}
\usepackage{comment}
\usepackage{booktabs}

\usepackage{makecell}
\usepackage{bm}
\usepackage{float}
\usepackage{afterpage}

\usepackage{multirow}
\usepackage[normalem]{ulem}
\usepackage[pagebackref=true,breaklinks=true,letterpaper=true,colorlinks,bookmarks=false]{hyperref}

\iccvfinalcopy 

\def\iccvPaperID{****} 

\begin{document}

\title{Urban Radiance Field Representation with Deformable Neural Mesh Primitives}

\author{Fan Lu\textsuperscript{1 *}~~~Yan Xu\textsuperscript{2 *}~~~Guang Chen\textsuperscript{1 \dag}~~~Hongsheng Li\textsuperscript{2,3,4}~~~Kwan-Yee Lin\textsuperscript{2,3 \dag}~~~Changjun Jiang\textsuperscript{1}\\
\textsuperscript{1}Tongji University~~~\textsuperscript{2}The Chinese University of Hong Kong~~~\textsuperscript{3}Shanghai AI Laboratory~~~\textsuperscript{4}CPII\\
{\tt\small\{lufan,guangchen,cjjiang\}@tongji.edu.cn} \\
\tt\small{yanxu@link.cuhk.edu.hk hsli@ee.cuhk.edu.hk junyilin@cuhk.edu.hk}
}

\twocolumn[
{%
\maketitle
\vspace{-9ex}
\begin{figure}[H]
\hsize=\textwidth 
\centering
\setlength{\abovecaptionskip}{1ex}
 \includegraphics[width=1\textwidth]{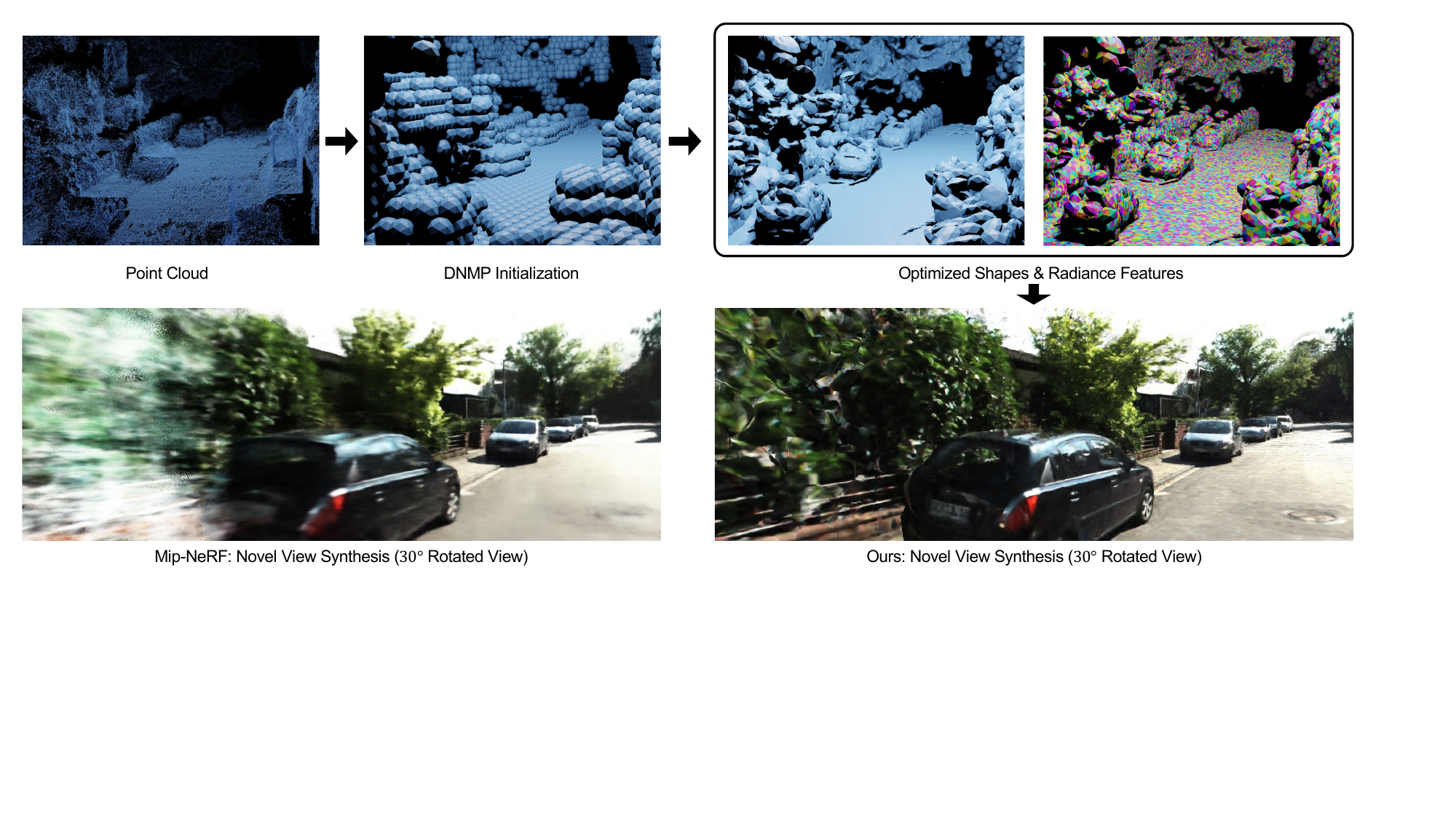}
\caption{
The basic idea of our method. Taking the patchy point clouds as inputs, we first voxelize the points and then initialize our Deformable Neural Mesh Primitive (DNMP) for each voxel. During training, the shapes of DNMPs are deformed 
to model the underlying 3D structures, while the radiance features of DNMPs are learnt to model the local radiance information for neural rendering. 
Based on our representation, we achieve efficient and photo-realistic rendering for urban scenes.
}\label{fig:teaser}
\end{figure}
\vspace{-2ex}
}
]
\ificcvfinal\thispagestyle{empty}\fi

{
\let\thefootnote\relax\footnotetext{* equal contribution. Fan's contributions: 1) Code implementation; 2) Novel database construction for latent space training and Loss Eq.~(1); 3) Conducted most of the experiments. 
Yan's  contributions: 1) Proposed DNMP. The structure design of DNMP and 
latent space construction for shape control; 
2) Proposed representing the entire scene with local primitives as well as the idea of hierarchical DNMP representation. 3) Drafted the paper and designed most of the experiments. }

\let\thefootnote\relax\footnotetext{\dag corresponding authors.}
}

\begin{abstract}
   Neural Radiance Fields (NeRFs) have achieved great success in the past few years. However, most current methods still require intensive resources due to ray marching-based rendering. To construct urban-level radiance fields efficiently, we design Deformable Neural Mesh Primitive~(DNMP), and propose to parameterize the entire scene with such primitives. The DNMP is a flexible and compact neural variant of classic mesh representation, which enjoys both the efficiency of rasterization-based rendering and the powerful neural representation capability for photo-realistic image synthesis. Specifically, a DNMP consists of a set of connected deformable mesh vertices with paired vertex features to parameterize the geometry and radiance information of a local area. To constrain the degree of freedom for optimization and lower the storage budgets, we enforce the shape of each primitive to be decoded from a relatively low-dimensional latent space. The rendering colors are decoded from the vertex features (interpolated with rasterization) by a view-dependent MLP. The DNMP provides a new paradigm for urban-level scene representation with appealing properties: $(1)$ High-quality rendering. Our method achieves leading performance for novel view synthesis in urban scenarios. $(2)$ Low computational costs. Our representation enables fast rendering (2.07ms/1k pixels) and low peak memory usage (110MB/1k pixels). We also present a lightweight version that can run 33$\times$ faster than vanilla NeRFs, and comparable to the highly-optimized Instant-NGP (0.61 vs 0.71ms/1k pixels). Project page: \href{https://dnmp.github.io/}{https://dnmp.github.io/}.
\end{abstract}

\section{Introduction}\label{sec:intro}

Synthesizing photo-realistic images of 3D scenes is a long-standing problem in computer vision, 
and has been the focus of research in the past decades.  
However, even with years of effort, the current paradigms still face great challenges especially in urban outdoor scenarios, due to the increased representation complexity and demanding computational resources.

To achieve high-quality image rendering,  the computer graphics community has explored various techniques for scene representation, including point clouds~\cite{kobbelt2004survey}, meshes~\cite{hoppe1993mesh,kazhdan2006poisson}, voxels~\cite{schwarz2010fast}, implicit functions~\cite{1467575,izadi2011kinectfusion}, \etc. 
The mesh-based representation is widely used in modern rendering pipelines due to its compact and efficient nature. However, constructing water-tight mesh models 
of urban scenes for modern graphic engines is still difficult. 
Besides, the textures and illuminations are difficult to be realistically recovered with the classic techniques. 

The recent neural rendering methods circumvent the mesh construction step and represent the scene with implicit neural functions. NeRF~\cite{mildenhall2021nerf} and its advanced variants~\cite{martin2021nerf,pumarola2021d,wang2021nerf,zhang2020nerfpp,barron2021mip} proposed to store the density and radiance information of a volume inside multi-layer perceptrons (MLPs), and adopt volumetric rendering for view synthesis. Recently,  some researchers~\cite{tancik2022block} also made efforts to extend the NeRF models to large-scale scenes by independently representing a city block-by-block and merging the representations together thereafter.   
Although remarkable progress has been made, their rendering process is still computationally intensive as the implicit functions need to be evaluated thousands of times to densely sample the space during volumetric rendering. 
Most of the computational resources are wasted on
the samples in empty spaces, and the situation will further escalate in outdoor scenes where empty space dominates.  

More recently, researchers have identified this issue and propose to combine neural rendering with explicit point-cloud reconstruction to improve the efficiency~\cite{xu2022point,ost2022neural}. In this way, the empty spaces can be skipped by taking the explicit reconstruction as a reference, which significantly saves computational resources. These methods associate point-wise learnable high-dimensional features to the reconstructed point clouds for spatial radiance encoding. During rendering, only the points around the intersections of the view ray with the point clouds will be sampled for feature aggregation. Based on such aggregation mechanism,  a dense and perfect reconstruction is vital for photo-realistic rendering. However, the reconstructed points are usually not uniformly distributed from the current reconstruction algorithms~\cite{goesele2006multi} and missing regions may be also ubiquitous. The noisy reconstruction will increase the learning difficulty of the implicit function and degrade the final rendering quality.

In this work, we propose an efficient radiance field representation for large-scale environments by combining efficient mesh-based rendering and powerful neural representations.    
Specifically, we develop Deformable Neural Mesh Primitive (DNMP) and propose to represent the entire radiance field in a bottom-up manner with such primitives, where each DNMP parameterizes the geometry and radiance of a local area. 
A DNMP consists of a set of connected deformable mesh vertices and each vertex is paired with a feature vector for radiance modeling. 
To constrain the degree of freedom for shape optimization and decrease the storage budgets, we enforce the mesh vertices of each DNMP to be decoded from a relatively low-dimensional latent code.  
The latent code will be optimized to deform the primitive shapes for 3D structure modeling during training.

Different from previous methods~\cite{xu2022point,ost2022neural} that rely on 
inefficient k-Nearest Neighbors (k-NN) algorithm to gather related features for rendering, we can directly leverage the rasterization pipeline for feature interpolation, which is {more} efficient.  
Based on rasterization, for each view ray, a set of features is collected by interpolations from the triangulated vertex features. 
These interpolated features are thereafter input to an implicit function (implemented with MLPs) to get the corresponding radiance and opacity values for 
volumetric rendering.  

To represent the entire scene, we first coarsely voxelize the scene 
according to the 3D reconstruction results (from Multi-View Stereo (MVS)~\cite{schoenberger2016mvs} or hardware sensors), and then parameterize the geometry and radiance of each voxel with a DNMP. 
Considering the practical 3D reconstruction results may be noisy and full of missing regions,   
we further propose to voxelize the scene with hierarchical resolutions   
and separately represent the radiance fields   
accordingly with hierarchically-sized DNMPs. 
The rendering results from different hierarchy levels will be blended.
Based on our DNMP-based hierarchical representation, we achieve more robustness against noisy 3D reconstructions compared with the previous point-cloud based methods~\cite{xu2022point,ost2022neural}.        

We evaluate our method on two urban datasets, \ie, KITTI-360~\cite{liao2022kitti} and Waymo Open Dataset~\cite{sun2020scalability}. 
Our method enables photo-realistic rendering and achieves 
leading performance for novel view synthesis.  
We achieve a much faster speed and produce fewer peak memory footprints compared with vanilla NeRFs. 
We also present a lightweight version to further accelerate the rendering.   
This lightweight version can run at an interactive rate only with limited sacrifices on rendering quality, the speed of which is even comparable with the highly-optimized Instant-NGP's~\cite{mueller2022instant}. 
Moreover, our method can be easily embedded into 
modern graphic rendering pipelines and 
naturally supports scene editing, which provides the potential for possible applications such as VR/AR.

\section{Related Work}
\noindent\textbf{Neural rendering.}
The early-phase neural rendering techniques~\cite{meshry2019neural,riegler2021stable,li2022read,aliev2020neural} proposed to directly project 3D signals to a 2D image plane and 
train a 2D CNN that maps projected signals to the final output image. Their mapping process only relies on the CNN regression ability without explicit physical modeling of the 3D space,  which could bring performance bottlenecks in synthesizing novel views. 
The recent volume rendering based approaches~\cite{mildenhall2021nerf,martin2021nerf,pumarola2021d,wang2021nerf,zhang2020nerfpp,barron2021mip,deng2022depth,rematas2022urban,fu2022panoptic} 
alleviate such dilemmas by storing the densities and radiances of a scene within an implicit neural function and synthesizing novel views with volumetric rendering.
Mip-NeRFs~\cite{barron2021mip,barron2022mip} render anti-aliased conical frustums instead of rays, which are widely applied
given the good rendering results in general. 
However, 
the implicit function needs to be evaluated thousands of times on the densely sampled space points for volumetric sampling, leading to inefficient training and inference. 
In the past few years, many methods are proposed to accelerate NeRFs. 
Some methods~\cite{hedman2021baking,liu2020neural,neff2021donerf,piala2021terminerf,xu2022point} use proxy scene structures or surface information 
to reduce the number of samples in empty spaces for inference acceleration,
while other solutions improve the inference or training speed by marrying NeRF with efficient data structures
~\cite{reiser2021kilonerf,garbin2021fastnerf,hedman2021baking,yu2021plenoctrees,sun2022direct,fridovich2022plenoxels,mueller2022instant,chen2022mobilenerf}.
As a representative, Instant-NGP~\cite{mueller2022instant} significantly improves the efficiency of classic NeRF via the hierarchical space division and highly-optimized CUDA implementation. 
However, these methods lack explicit surface constraints, which may lead to less robustness against viewpoint changes for view synthesis.     
Moreover, all these methods still have not been widely   supported by modern graphic rendering pipelines and also have difficulties in supporting scene editing for downstream applications. 

\noindent\textbf{Outdoor NeRFs.}
Recently, some researchers tried to extend NeRF to outdoor scenes~\cite{rematas2022urban,martin2021nerf,tancik2022block,ost2022neural}. 
NeRF-W~\cite{martin2021nerf} incorporates frame-specific codes in the rendering pipeline to handle the photometric variation and transient objects. Block-NeRF~\cite{tancik2022block} extends the NeRF to urban scenarios and builds up the block-wise radiance fields with individual NeRFs to composite the complete scene. 
However, these methods still rely on costly volumetric sampling as indicated above, which will waste huge amounts of computational resources in empty spaces. 
Rematas~\etal~\cite{rematas2022urban} include LiDAR point clouds for supervision to facilitate geometry learning.
Neural Point Light Field (NPLF)~\cite{ost2022neural} leverages the explicit 3D reconstructions from LiDAR data to represent the radiance field for rendering efficiency. 
But they still simply aggregate several nearest feature points around each view ray for rendering without considering the scene geometry in detail, which causes bottlenecks for high-resolution rendering.    

\noindent\textbf{3D Shape Reconstruction}. 
The classic pipeline for 3D reconstruction from color images usually first estimates the camera poses based on structure-from-motion~\cite{schoenberger2016sfm, agarwal2011building} and then recovers dense depths with Multi-View Stereo (MVS) techniques~\cite{schoenberger2016mvs,cheng2020deep,furukawa2009accurate,yao2018mvsnet,wang2021patchmatchnet}. These methods can handle ideal scenarios but may generate incomplete reconstruction results under adverse conditions, \eg, illumination changes, textureless areas, \etc. 
The methods based on implicit fields \cite{peng2020convolutional,duan2020curriculum,Yu2022MonoSDF,wang2021neus} are generally more robust but an expensive iso-surfacing step~\cite{lorensen1987marching} is required to extract the mesh from the representation, which is prone to quantization errors. 
Prior to our work, some works \cite{wang2018pixel2mesh,gao2020learning,shen2021deep,munkberg2022extracting} 
proposed deformable meshes for shape reconstruction and rendering. However, these techniques are still sensitive to practical noisy data and are usually limited to object-level shape optimization. 

\section{Method}
To parameterize the large-scale urban radiance field effectively, we propose Deformable Neural Mesh Primitive (DNMP). 
DNMP is a neural variant of classic mesh representation, which takes advantage of both the efficient rasterization-based rendering and the powerful neural representation capability.  
A DNMP is capable of modeling the geometry and radiance information of a local 3D space in an expressive and compact manner, and the entire radiance field is hierarchically constituted by a series of DNMPs.  

\subsection{Deformable Neural Mesh Primitive}\label{sec:DNMP}
Triangle meshes are widely used in computer graphics. The meshes can compactly represent 3D surfaces and be efficiently rendered based on rasterization. 
To leverage the efficiency of mesh-based representation and the impressive radiance representation ability of neural features,   
we develop Deformable Neural Mesh Primitive (DNMP) and parameterize the entire scene with such primitives. 
DNMP is an enclosed triangle neural mesh, constituted by a set of deformable mesh vertices $\mathcal{V}=\{\mathbf{v}_i|i=1,\ldots,N\}$  paired with learnable vertex radiance features $\mathcal{F}=\{\mathbf{f}_i|i=1,\ldots, N\}$.  
The vertices define the shape of each DNMP, while the vertex features encode the radiance information of a local area.   

\noindent\textbf{Shape parameterization of DNMP.} 
A DNMP may contain tens of vertices. 
To constrain the degree of freedom for shape optimization, 
we design an auto-encoder~\cite{kramer1991nonlinear} to learn a compact latent space to parameterize the primitive shapes.

Specifically, we first establish a database that contains a huge number of meshes created from local 3D structures\footnote{The details are in supplementary materials.} and train our auto-encoder with them. These local structures are collected from different types of datasets, both indoors and outdoors.      
We assume the database is large enough to capture all the possible local structural variations for compact latent space learning.  
\begin{figure*}[t!]
    \centering
    \setlength{\abovecaptionskip}{0.5ex}
    \includegraphics[width=1\linewidth, trim=0cm 7.5cm 4cm 0cm, clip]{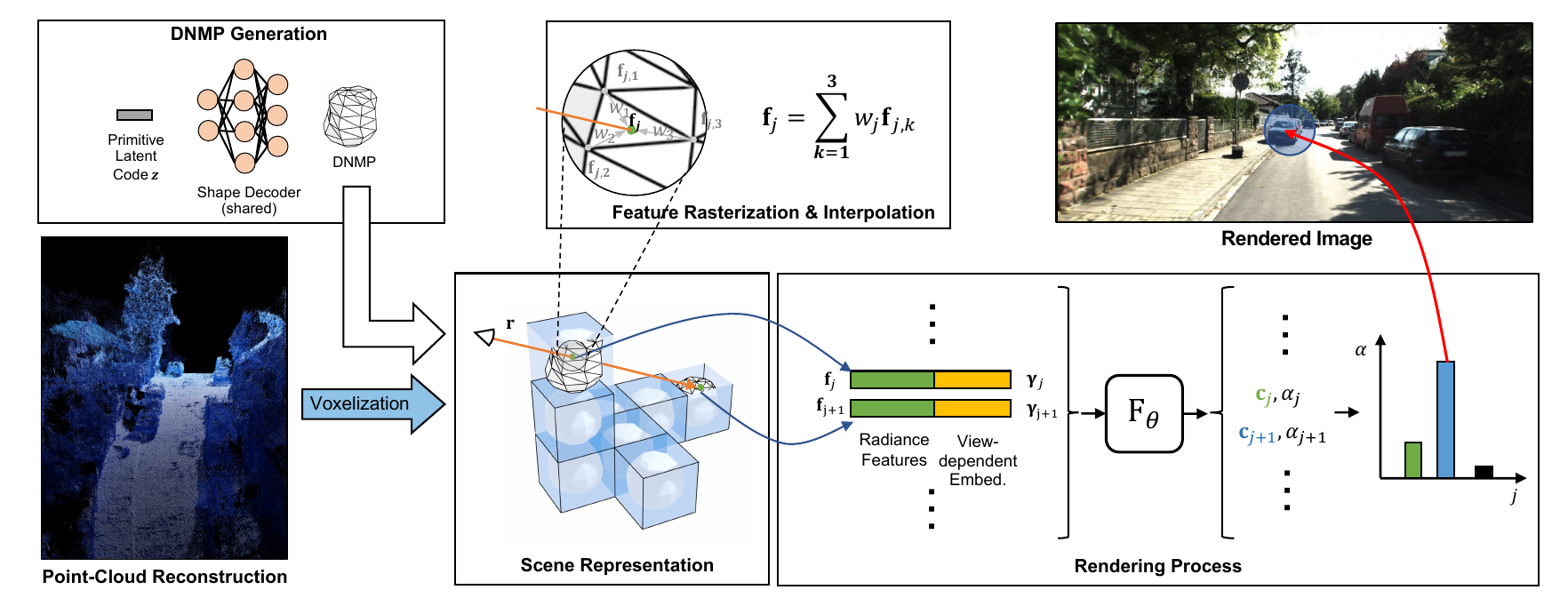}
    \caption{ 
    The overview of our framework.
    The entire scene is voxelized based on the point-cloud reconstruction, where each voxel is assigned a DNMP to parameterize the geometry and radiance of the local area.  
    By rasterization, we can obtain the interpolated radiance features $\{\mathbf{f}_j|j=0,1,\ldots\,J\}$ from the intersected DNMPs for each view ray $\mathbf{r}$.     
    Thereafter, these interpolated features along with the view-dependent 
    embeddings $\{\bm{\gamma}_j| j=0,1,\ldots,J \}$ are sent to an implicit function $\mathrm{F}_\theta$ to predict the radiance value $c_j$ and opacity $\alpha_j$ of each intersection point. Finally, the rendering color $\hat{\mathbf{C}}(\mathbf{r})$ of the view ray $\mathbf{r}$ is obtained by blending the radiance values according to the opacities  $\{\alpha_j | j=0,1,\ldots, J\}$\protect\footnotemark[2].
    }
    \label{fig:framework}
    \vspace{-4ex}
\end{figure*}

Our auto-encoder is designed based on PointNet~\cite{qi2017pointnet}, containing a shape encoder and a shape decoder. 
The shape encoder  
encodes the geometric information of differently-shaped meshes into compact shape latent codes $\mathbf{z}$. The shape decoder $G$ 
directly decodes the shape latent code $\mathbf{z}$ to DNMP's vertices $\mathcal{V}=\{\mathbf{v}_i|i=1,\ldots,N\}$ (in a predefined order), \ie, $\mathcal{V} = G(\mathbf{z})$.
The output vertices can be trivially converted to an enclosed mesh according to the predefined connectivity relations.
To learn the latent space, we encourage the shapes of decoded DNMPs to match the input meshes.
The training loss of the auto-encoder contains two parts:
\begin{equation}
   L_{ae} =  L_{chamfer}(\mathcal{V} ) + L_{regularize}(\mathcal{V}).  
\end{equation}

The chamfer distance loss $L_{chamfer}$ enforces the closeness between two sets of randomly sampled points from the input and output mesh faces.     
Besides, we also encourage the normal consistency~\cite{desbrun1999implicit} and mesh smoothness~\cite{nealen2006laplacian} with a regularization loss $L_{regularize}$, which is described detailedly in supplementary materials.
To constrain the latent code space, we normalize the latent code to unit length, \ie, $||\mathbf{z}||_2=1$. The dimension of the latent space is empirically set to $8$ for reliable shape optimization.

\noindent\textbf{Radiance parameterization. }
To encode the radiance information, we associate independent learnable feature vectors to DNMP's vertices. During rendering, unlike the previous point-based methods~\cite{xu2022point,ost2022neural} that need to query features with time-consuming k-NN searching~\cite{fix1989discriminatory}, we can efficiently obtain the related features via rasterization. 
Moreover, in this manner, the encoded radiance is better aligned with the local surfaces,   
which is helpful to improve the view consistency of the radiance field.      

\footnotetext[2]{We only show the rendering process for one hierarchy in this figure. The final color will be blended from several hierarchies with Eq.~\ref{eq:multi_scale_blending}.}

\subsection{DNMP-Based Scene Representation}\label{sec:geo_reconstruct}
To construct a view-consistent radiance field, we emphasize the optimization of both geometry and radiance. 
For initialization, the target scene is first voxelized based on the point-cloud reconstruction from MVS~\cite{schoenberger2016mvs} or hardware sensors. 
Then, each voxel is assigned a DNMP to parameterize the structure geometry and radiance information in the local area.  
Before training, the shape latent code of each DNMP is initialized to correspond to a spherical template (scaled to match the voxel size), as illustrated in Fig.~\ref{fig:teaser}. For convenience, we denote the set of all the shape latent codes of the scene as 
$\mathcal{Z}=\{\mathbf{z}_l| l=1,2,\ldots,L \}$, where $l$ indexes the DNMP and $L$ is the total number of DNMPs of the scene.

\begin{figure}
\setlength{\abovecaptionskip}{0.5ex}
\centering
\resizebox{0.9\linewidth}{!}{
 \includegraphics[angle=0,origin=c, width=1\linewidth, trim=0cm 0cm 12cm 0cm, clip]{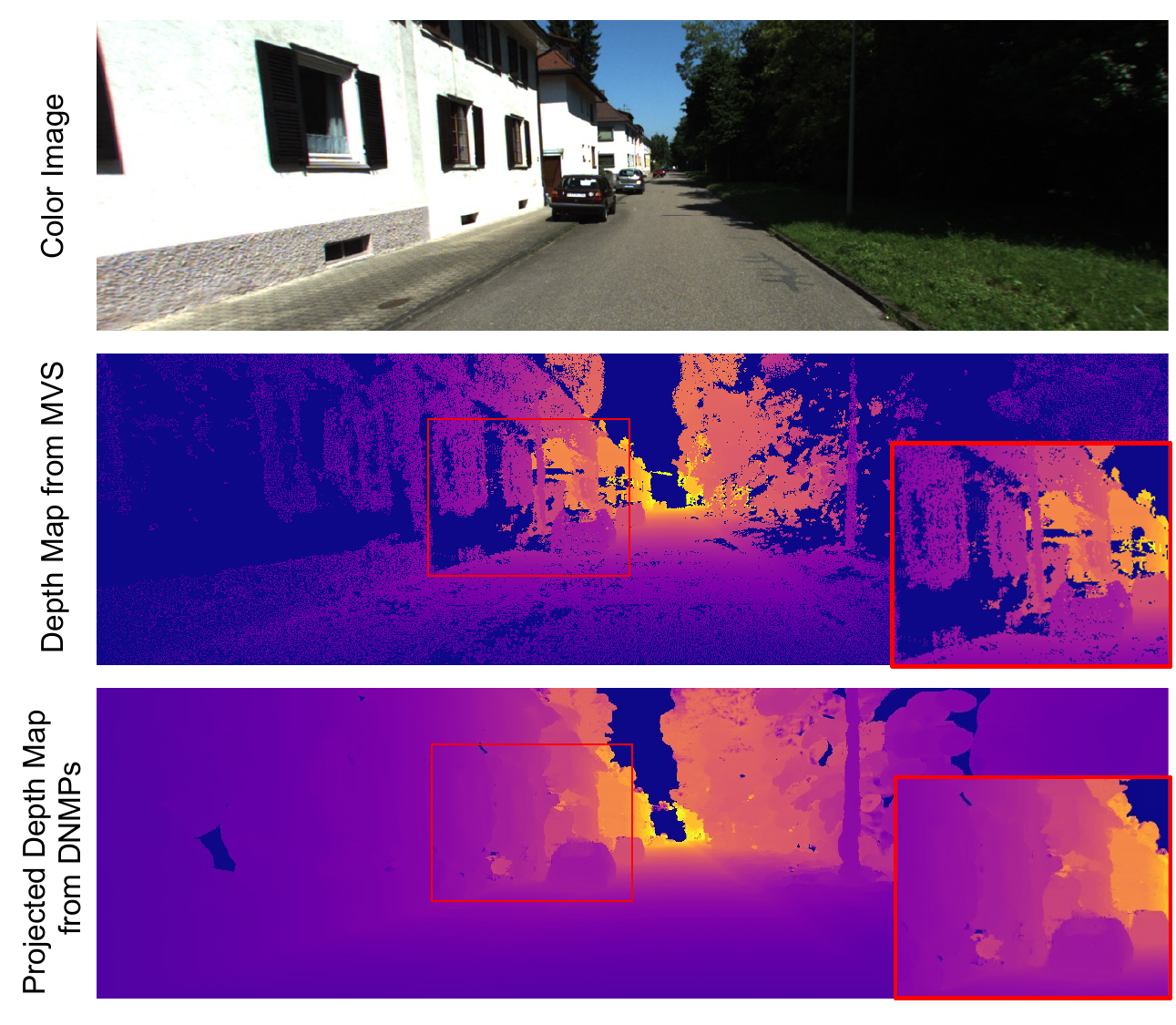}
 }
\caption{Even supervised with the incomplete depths with Eq.~\eqref{eq:loss_geo}, the DNMPs can be generally deformed to reflect the underlying geometry thanks to the compact latent space learnt from huge number of local 3D structures.}\label{fig:depth_completion}
\vspace{-3ex}
\end{figure}
\noindent\textbf{DNMP-based shape optimization.}
To abstract the scene's geometry with DNMPs, we need to optimize the shape latent codes.
We leverage 
the estimated depth maps $D$ (by MVS~\cite{schoenberger2016mvs} or other sensors) of captured video sequences of the scene to optimize the latent code.    
During training, we render a corresponding depth map $\hat{D}$ based on the current shapes of DNMPs in a differentiable manner.  As the DNMP shapes are controlled by the latent codes $\mathcal{Z}$,  the rendered depth $\hat{D}$ can be written as a differentiable function of latent codes $\mathcal{Z}$ as  $\hat{D}(\mathcal{Z})$. 
Then, we supervise the rendered depth map $\hat{D}$ with the pre-estimated depth map $D$ via the L1 loss 
\begin{equation}\label{eq:loss_geo}
    L_{geo} = || \hat{D}( \mathcal{Z} )-D  ||_1,   
\end{equation}
which deforms the DNMPs to abstract the scene's geometry.  
We found that even supervised with imperfect depth maps $D$ (shown in Fig.~\ref{fig:depth_completion}), the latent codes $\mathcal{Z}$ can be optimized to reflect the underlying geometry reasonably, thanks to the strong structural priors learnt by the pretrained shape decoder~(Sec.~\ref{sec:DNMP}) from huge amounts of data.

\noindent\textbf{Hierarchical representation.}\label{sec:hierarchical_rep}
As the 3D reconstruction results with MVS~\cite{schoenberger2016mvs} may have many missing regions in outdoor environments,  DNMP may thus fail to be initialized for the missing parts, which will cause unsatisfactory rendering results. 
To avoid this, we model the entire scene in a hierarchical manner. 
Concretely, we voxelize the reconstructed point cloud with  hierarchical sizes, \eg, 0.5m, 1m, \etc. to make the missing regions can be covered by larger-sized voxels.  
Thereafter, we initialize the DNMP (with respectively scaled sizes) for each hierarchy  and optimize their shapes with Eq.~\eqref{eq:loss_geo}.   

\subsection{Radiance Modeling and View Synthesis}\label{sec:rendering}
\noindent\textbf{Rasterization and radiance feature interpolation.} 
To render a pixel, we collect the related features with rasterization as shown in Fig.~\ref{fig:framework}. 
The radiance features used for rendering, denoted as $\{\mathbf{f}_j|j=1,2\ldots, J\}$, 
are essentially interpolated from the triangulated vertex features of the intersected mesh faces, where $j$ indexes the intersections.   
Besides, we use the view-ray direction  $\mathbf{d}\in \mathbb{R}^3$  and the surface normal $\mathbf{n}_j\in\mathbb{R}^3$ at the intersection point to model the view-dependent factor $\bm{\gamma }_j=\{\mathbf{n}_j,\mathbf{d}\}$. 

\noindent\textbf{Rendering.}
The radiance feature $\mathbf{f}_j$ as well as view-dependent factor $\bm{\gamma}_j$ is input to a MLP $\mathrm{F}_\theta$ (shared among a sequence) after positional encoding~\cite{mildenhall2021nerf}. $\mathrm{F}_\theta$ will predict a pair of radiance value $\mathbf{c}_j$ and opacity $\alpha_j$ for each intersected point:
\begin{equation}
    \mathbf{c}_j, \alpha_j = \mathrm{F}_\theta(\mathbf{f}_j, \bm{\gamma}_j). 
\end{equation}
The opacity $\alpha_j$ here represents the probability that the ray will terminate at the $j$-th point \cite{attal2022learning}. 
In our implementation, $\mathrm{F}_\theta$ predicts the opacity $\alpha_j$ only based on $\mathbf{f}_j$. The view-dependent factor $\bm{\gamma}_j$ is only input to the branch split from near the end of the entire network to predict the view-dependent radiance value $\mathbf{c}_j$ following the spirit of \cite{mildenhall2021nerf}.

We empirically keep $J$ nearest intersections for each view ray and predict their radiance and opacity values, \ie, $\{ (\mathbf{c}_j, \alpha_j)| j=1,\ldots,J \}$. Thereafter, to render the pixel of this view ray $\mathbf{r}$, the expected color is calculated as 
\begin{equation}\label{eq:volume_rendering}
    \hat{\mathbf{C}}(\mathbf{r}) = \sum_j^J T_j\alpha_j\mathbf{c}_j, ~~~~T_j = \prod_{p=1}^{j-1}(1-\alpha_p). 
\end{equation}

\noindent\textbf{Blending hierarchical DNMPs.} 
To improve the unsatisfactory rendering caused by the missing regions in point-cloud reconstruction, we need to blend the rendering results of DNMPs from different hierarchies. We denote the rendered colors from different hierarchies with Eq.~\eqref{eq:volume_rendering} as $\{\hat{\mathbf{C}}_1(\mathbf{r}), \hat{\mathbf{C}}_2(\mathbf{r}), \ldots, \hat{\mathbf{C}}_S(\mathbf{r})\}$, where the subscripts denote the hierarchy levels.       
We practically blend these results from the finest level ($\hat{\mathbf{C}}_1(\mathbf{r})$) to the coarsest level ($\hat{\mathbf{C}}_S(\mathbf{r})$) to better keep the texture details:    
\begin{equation}
    \label{eq:multi_scale_blending}
    \hat{\mathbf{C}}(\mathbf{r})=\hat{\mathbf{C}}_1(\mathbf{r})+(1-\mathcal{A}_1)\hat{\mathbf{C}}_2(\mathbf{r})+\cdots+\prod_{s=1}^{S-1}(1-\mathcal{A}_s)\hat{\mathbf{C}}_S(\mathbf{r}), 
\end{equation}
where $\mathcal{A}_s$ is calculated as the summation of the accumulated weights of the respective hierarchy, \ie, $\mathcal{A}_s=\sum_j^J T_j\alpha_j$, and $\mathcal{A}_s$ will be manually set to 0 if there is no view-ray intersection in this level. 

\noindent\textbf{Radiance feature learning. }     
The DNMPs' vertex features are empirically initialized with the positional encoding~\cite{mildenhall2021nerf} of the vertex coordinates before training.   
During training, we supervise the radiance feature learning with the camera images $\mathbf{C}$ in the training video sequence: 
\begin{equation}\label{eq:radiance_loss}
    L_{rad} = \sum_{\mathbf{r}\in \mathcal{R}} ||\hat{\mathbf{C}}(\mathbf{r}) - \mathbf{C}(\mathbf{r}) ||_2^2, 
\end{equation}
where $\hat{\mathbf{C}}(\mathbf{r})$ and  $\mathbf{C}(\mathbf{r})$ are the rendered color and the ground-truth color of view ray $\mathbf{r}$.  This loss is applied for all the view rays $\mathcal{R}$ defined by the training image sequence.      

\noindent\textbf{Non-structured regions.} Given that our method mentioned above is based on the explicit geometric abstraction of the scene, the non-structured regions (\eg, sky) cannot be well handled. 
We use Mip-NeRF~\cite{barron2021mip} to handle these regions in our implementation. 

\section{Experiments}

\subsection{Experimental Setup}
\noindent\textbf{Implementation details.} For the hierarchical representation, we voxelize the point clouds with two hierarchical sizes, \emph{i.e.}, 0.5m and 1m. 
During rendering, we blend radiance values of the nearest 4 intersection points for the $0.5$m hierarchy, \ie, $J$ is set to $4$. For the hierarchy with $1$m DNMPs, $J$ is set to $2$. 
The frequency of positional encoding for vertex feature initialization is set to $3$, resulting in $21$-dimensional features. Please refer to supplementary materials for more details.

\noindent\textbf{Datasets.} We conduct experiments on two 
urban datasets, \emph{i.e.}, KITTI-360~\cite{liao2022kitti} and Waymo Open Dataset~\cite{sun2020scalability}. KITTI-360 is a large-scale dataset containing $4\times83,000$ images captured in urban environments with a driving distance of around $73.7$ km. 
We select $5$ sequences from them to evaluate our method.  
For the evaluation of novel view synthesis, we select every second image in each sequence as the test set and train our model on the remaining images.  
For Waymo Open Dataset, we follow~\cite{ost2022neural} to select the $6$ sequences mainly containing static objects for our experiments. 
We select every $10$th image in the sequences as the test set and take the remaining ones as the training set. 
Details of datasets splits are provided in supplementary materials.
To better evaluate and compare the synthesis capability for details, 
we train and evaluate our methods and all the baselines with full-resolution images 
(\ie, $1408\times 376$ for KITTI-360 and $1920\times 1280$ for Waymo dataset).

\noindent\textbf{Evaluation metrics.} Following the previous methods~\cite{mildenhall2021nerf,barron2021mip}, our evaluations are based on three widely-used metrics, \ie, peak signal-to-noise ratio (PSNR), structural similarity index measure (SSIM), and the learned perceptual image patch similarity (LPIPS)~\cite{zhang2018unreasonable}.

\subsection{Ablation Study}
We conduct thorough ablation studies on 
KITTI-360 dataset to analyze the effects of the proposed components on the task of novel view synthesis.

\begin{table}[t]
\centering
\setlength\tabcolsep{4pt} 
\renewcommand\arraystretch{1.15}
\caption{
Ablation studies to show the effectiveness of our DNMP-based shape optimization on KITTI-360 dataset.
}
\label{tab:shape_optim}

\resizebox{\linewidth}{!}{
\begin{tabular}{l c c c c c c c c c c c c c c c c c c c }
\toprule
\multirow{1}{*}{\textbf{Method}} & \multicolumn{1}{c}{direct shape optim.} & \multicolumn{1}{c}{w/o shape optim.} & \multicolumn{1}{c}{w/o hierarchy}  &\multicolumn{1}{c}{Ours}\\
\midrule
PSNR$\uparrow$ & 21.41 & 20.36 & 23.21 & \textbf{23.41} \\  
SSIM$\uparrow$ & 0.789 & 0.758 & 0.840 & \textbf{0.846} \\  
LPIPS$\downarrow$ & 0.397 & 0.421 & 0.322 & \textbf{0.305} \\  
\bottomrule
\end{tabular}
}
\vspace{-2ex}
\end{table}

\begin{figure}
\setlength{\abovecaptionskip}{1ex}
\resizebox{1.0\linewidth}{!}{
 \includegraphics[width=1\linewidth]{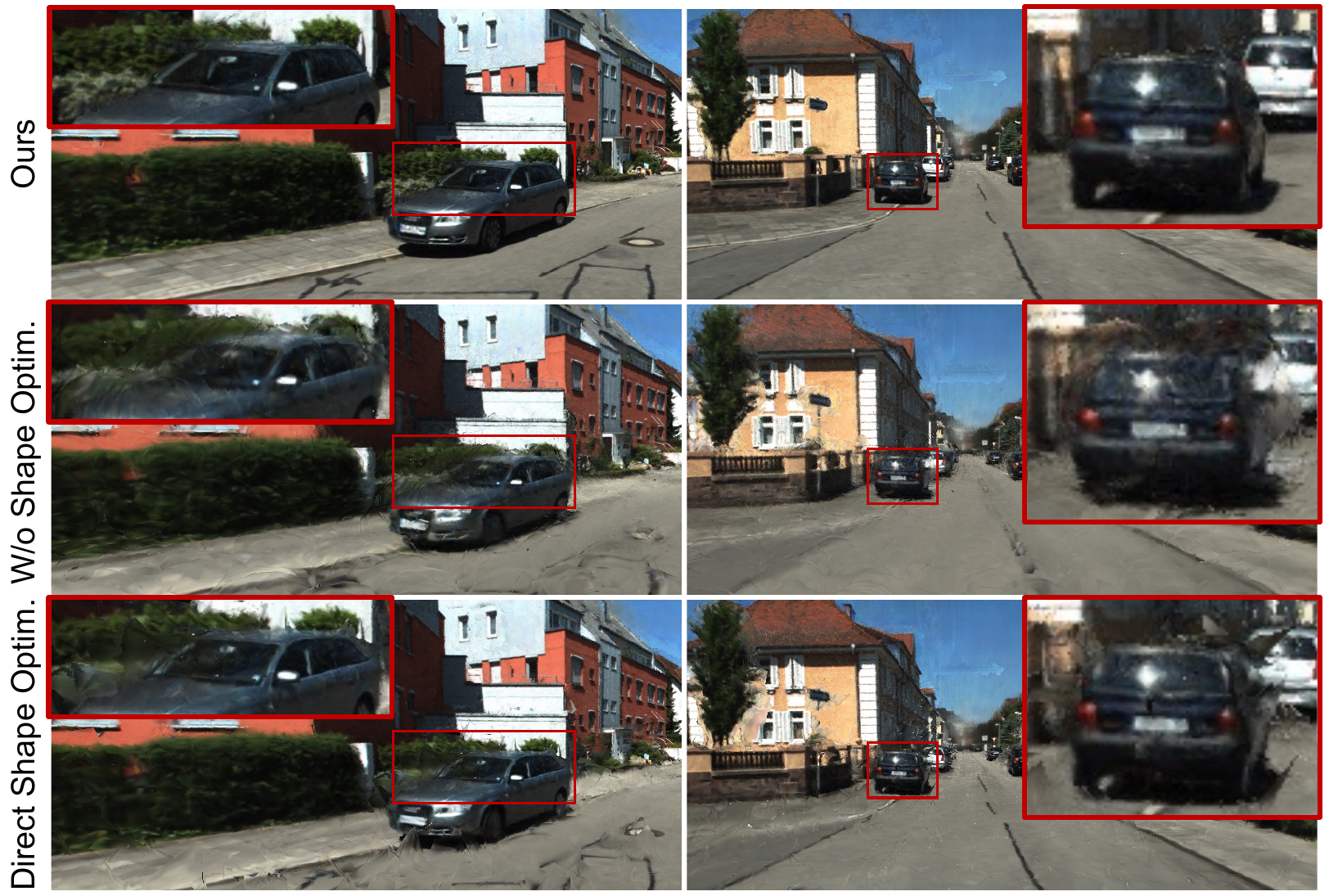}
 }
\caption{View synthesis results with different strategies for primitive shape optimization. The results of \textit{direct shape optim.} are slightly better than \textit{w/o shape optim.}, but are still much less satisfactory compared with our DNMP-based shape optimization. 
}
\label{fig:shape_optim}
\vspace{-3ex}
\end{figure}
\begin{figure}
\centering
\setlength{\abovecaptionskip}{1ex}
 \includegraphics[width=1.\linewidth, trim=0cm 6.5cm 3cm 0cm, clip]{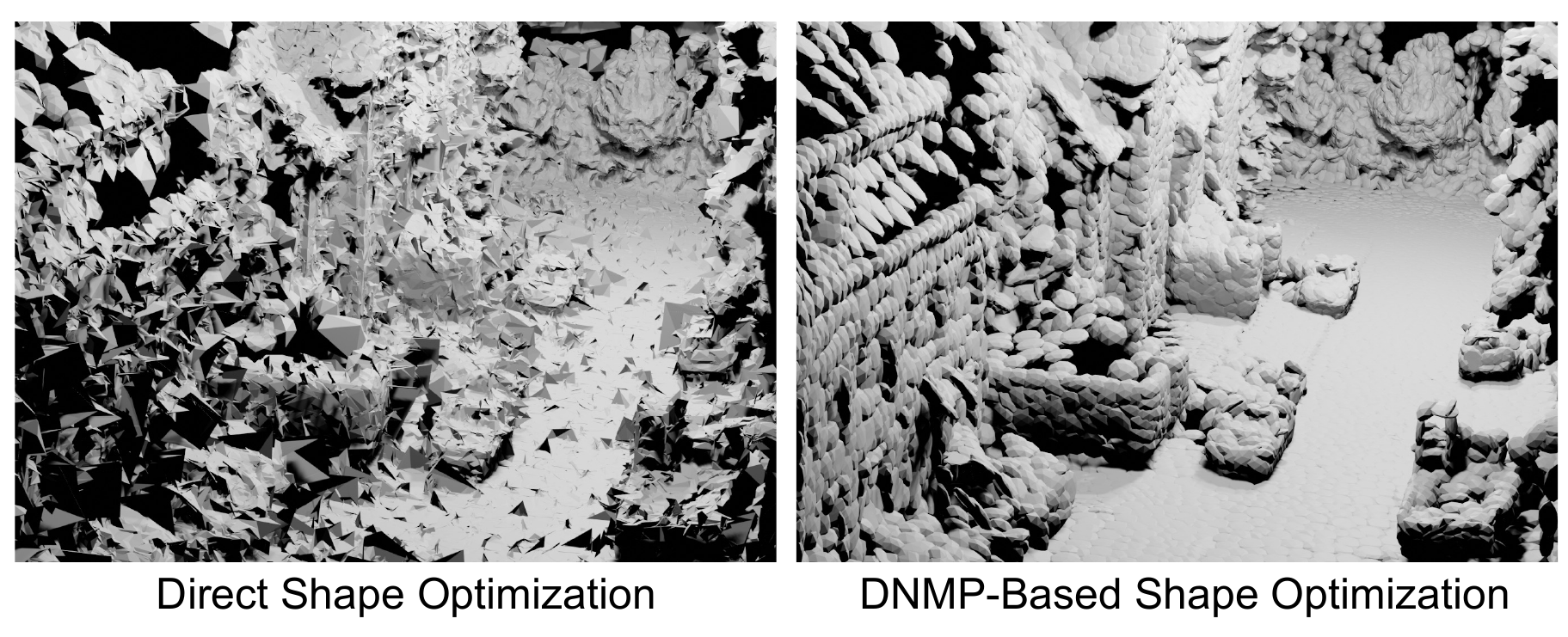}
\caption{Visualization of the scene geometry optimized with different strategies. Compared with directly optimizing the vertex parameters, our DNMP-based shape optimization is more robust. 
}
\label{fig:shape_optimization}
\vspace{-1ex}
\end{figure}

\noindent\textbf{Shape optimization of DNMPs.}
We introduce two ablation variants 
to demonstrate the effectiveness of the proposed DNMP-based shape optimization. For the variant \textit{direct shape optim.} in Table~\ref{tab:shape_optim}, instead of decoding the DNMP vertices from the latent code, we assign learnable vertex offset parameters to deform the spherical mesh templates and directly optimize these parameters using the same loss function $L_{geo}$ (Eq.~\eqref{eq:loss_geo}).  
For the variant \textit{w/o shape optim.}, we simply use the original spherical mesh templates without further shape optimization.  
According to the novel view synthesis performance in Table~\ref{tab:shape_optim} and visualizations in Fig.~\ref{fig:shape_optim}, although \textit{direct shape optim.} achieves better results than \textit{w/o shape optim.}, 
it is still inferior to our DNMP-based shape optimization. 
We visualize and compare the optimized scene geometries in Fig.~\ref{fig:shape_optimization}. 
Without the constraints on degree of freedom,  
the version \textit{direct shape optim.} becomes sensitive to incomplete and noisy depth maps from MVS, which leads to noisy geometry recovery. 
The unsatisfactory scene geometry should explain the degradation in rendering quality.   
In contrast, our proposed DNMP-based shape optimization is much more stable, which leads to 
robust surface modeling and better rendering quality.

\begin{table}[t]
\centering
\footnotesize
\setlength\tabcolsep{2pt}
\renewcommand\arraystretch{1.15}
\caption{
Hyperparameter analysis based on KITTI-360 dataset. 
}
\resizebox{\linewidth}{!}{
\begin{tabular}{l c c c c c c c c c c c c c c }
\toprule
\multirow{2}{*}{\textbf{Metric}} & \multicolumn{3}{c}{\# Intersection points $J$} && \multicolumn{3}{c}{Radiance feature dim.} && \multicolumn{3}{c}{DNMP radius} & \multirow{2}{*}{\makecell[c]{Light-\\weight}} \\
 \cmidrule{2-4} \cmidrule{6-8} \cmidrule{10-12}     
 & 2 & 4 & 8 && 15 & 21 & 27 && 2 m & 1 m & 0.5 m & \\  
\midrule
PSNR$\uparrow$ & 23.23  & 23.41 & \textbf{23.43} && 23.21 & \textbf{23.41} & 23.28 && 22.61 & 23.20 & \textbf{23.21} & 23.27 \\  
SSIM$\uparrow$ & 0.843 & 0.846 & \textbf{0.847} && 0.839 & \textbf{0.846} & 0.845 && 0.818 & 0.838 & \textbf{0.840} & 0.842   \\
LPIPS$\downarrow$ & 0.313 & \textbf{0.305} & 0.306 && 0.321 & 0.305 & \textbf{0.301} && 0.376 & 0.332 & \textbf{0.322} & 0.307 \\          
\bottomrule
\end{tabular}
}
\label{tab:ablation}
\vspace{-4ex}
\end{table}

\noindent\textbf{Rendering process.}
As we mentioned in Sec.~\ref{sec:rendering}, 
we collect the $J$ nearest radiance features from the mesh faces intersected by the view ray and calculate the rendering color with Eq.~\eqref{eq:volume_rendering}. 
We conduct experiments with different intersection numbers $J$ and the performance comparison is shown in  Table~\ref{tab:ablation} (in the column \textit{\# Intersection points} $J$). 
Our performance is relatively robust to the intersection number $J$ and a small $J=4$ is generally enough for good view synthesis performance due to our effective DNMP-based shape optimization.     
Besides, we also try  
with the radiance features of different dimensions (denoted as \textit{Radiance feature dim.}
in Table~\ref{tab:ablation}). It is shown that $21$-dimensional feature is sufficient for good rendering quality.  

\begin{figure}
\setlength{\abovecaptionskip}{1ex}
\centering
{
 \includegraphics[width=0.85\linewidth]{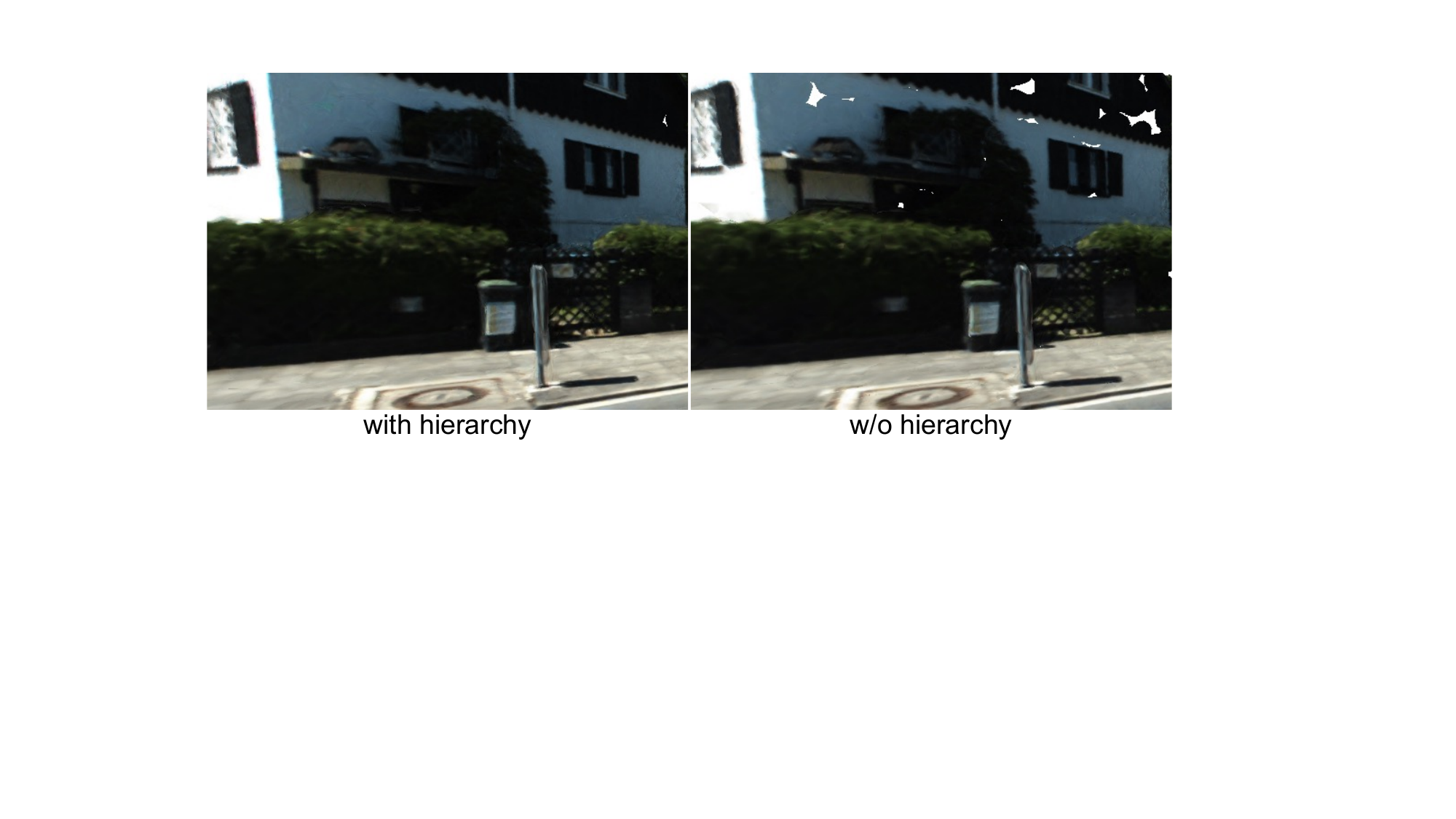}
 }
\caption{Rendering with or w/o hierarchical DNMPs. The proposed hierarchical DNMPs 
effectively completes the missing regions in the synthesized images. Better zoom in for more details.
}
\label{fig:hierarchy}
\vspace{-4ex}
\end{figure}
\noindent\textbf{Hierarchical DNMPs.}
We first analyze the performance with different radii of DNMPs. As shown in Table~\ref{tab:ablation} (columns under \textit{DNMP radius}), our performance is tolerant to large DNMP radii, which is beneficial for resource-constrained scenarios. To verify the necessity of the hierarchical representation in covering missing structures, we synthesize the images only with the DNMPs of the finest hierarchy, the performance of which is reported in Table~\ref{tab:shape_optim} (denoted as \textit{w/o hierarchy}).  The performance is degraded.
Examples shown in Fig.~\ref{fig:hierarchy} demonstrate that the hierarchical DNMPs can effectively improve the completeness of the synthesized images. 

\noindent\textbf{Lightweight version.} To better adapt to latency-sensitive systems, we provide a lightweight version by reducing the complexity of the MLP $\mathrm{F}_\theta$ (Please refer to supplementary materials for more details). Due to the high capacity of our surface-aligned features, the lightweight version still achieves competitive performance as shown in Table~\ref{tab:ablation}.

\subsection{Comparison with the State-of-the-Art Methods}
\begin{figure*}[t!]
\setlength{\abovecaptionskip}{1ex}
    \centering
    \includegraphics[width=1.\linewidth, trim=0cm 0cm 0cm 0cm, clip]{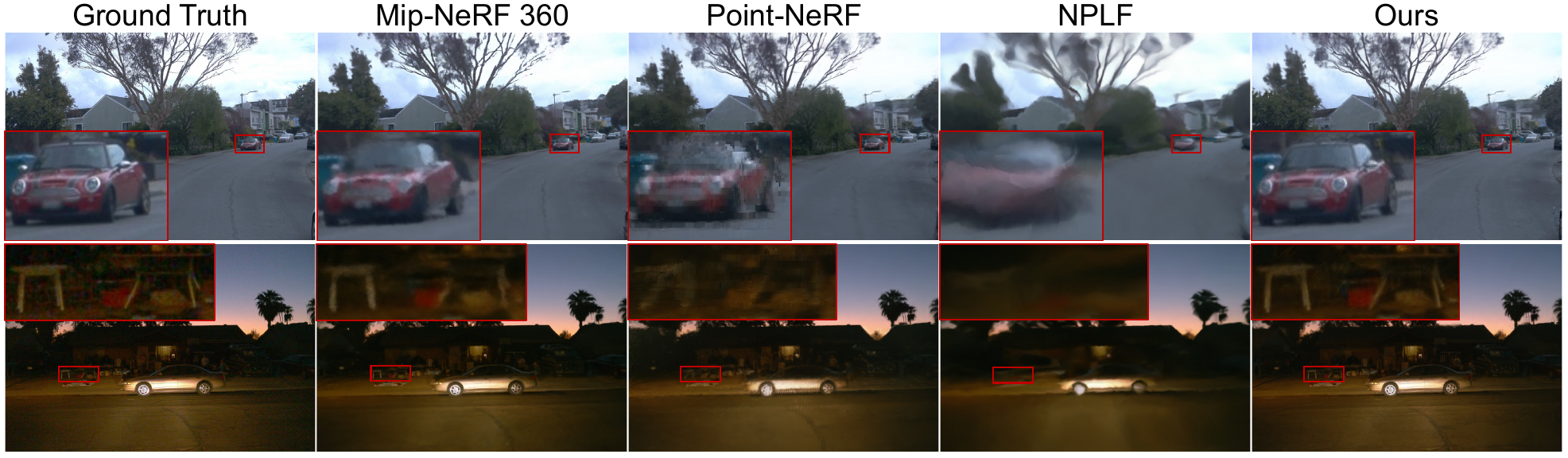}
    \caption{Qualitative comparison of novel view synthesis on Waymo dataset. Cropped patches are scaled to highlight the details. Due to the explicit and accurate surface modeling, the proposed method significantly outperforms baseline methods and effectively recover the texture details. Please refer to our supplementary materials for more visualization results.
    }
    \label{fig:comparison_waymo}
    \vspace{-3ex}
\end{figure*}
We compare the proposed method with several competitive baselines on the task of novel view synthesis, including NeRF~\cite{mildenhall2021nerf}, 
NeRF-W~\cite{martin2021nerf},
Mip-NeRF~\cite{barron2021mip},  Mip-NeRF 360~\cite{barron2022mip}, Instant-NGP~\cite{mueller2022instant}, and two point cloud-based neural rendering methods (\ie, Point-NeRF~\cite{xu2022point} and NPLF
(Neural Point Light Fields)~\cite{ost2022neural}).
The evaluation results on KITTI-360 dataset and Waymo dataset are shown in Table~\ref{tab:kitti},~\ref{tab:waymo}, respectively.  
We also provide qualitative comparisons on Waymo dataset in Fig.~\ref{fig:comparison_waymo} (Please refer to our supplementary materials for more visualization results). 
It is shown that our method performs consistently well on both datasets. Compared with these competitive baselines, our method achieves competitive novel view synthesis quality in terms of all the evaluation metrics. Especially, we achieve a lower LPIPS than other baselines. The metric LPIPS should be more consistent with the perception of our human beings~\cite{zhang2018unreasonable}. 
The lower LPIPS indicates that our synthesized images should be more realistic and rich in details compared with the other methods.

We also find that,  
although Point-NeRF has achieved good performance in small-scale synthetic dataset,   
it may fail to render high-quality images in real outdoor environments due to the existence of complex occlusions and incomplete noisy point-cloud reconstructions. 
Moreover, the previous neural rendering method NPLF designed for outdoor scenarios is also found to be less effective to synthesize high-resolution images. We deem it could be caused by the lack of geometry optimization in NPLF.
Thanks to the DNMP-based scene representation, the proposed method achieves photo-realistic rendering with rich  texture details.

\begin{table}[t]
\centering
\setlength\tabcolsep{4pt} 
\renewcommand\arraystretch{1.15}
\footnotesize
\caption{Performance comparison of novel view synthesis with other competitive baselines on KITTI-360 dataset.}
\begin{tabular}{l c c c}
\toprule
\textbf{Method} & PSNR$\uparrow$ & SSIM$\uparrow$ & LPIPS$\downarrow$ \\
\midrule
NeRF~\cite{mildenhall2021nerf} & 21.94 & 0.781 & 0.449 \\
NeRF-W~\cite{martin2021nerf} & 22.77 & 0.794 & 0.446 \\
Instant-NGP~\cite{mueller2022instant} & 22.89 & 0.836 & 0.353 \\
Point-NeRF~\cite{xu2022point} & 21.54 & 0.793 & 0.406 \\ 
Mip-NeRF~\cite{barron2021mip} & 23.21 & 0.810 & 0.455 \\
Mip-NeRF 360~\cite{barron2022mip} & \underline{23.27} & \underline{0.836} & \underline{0.355} \\
\midrule
Ours & \textbf{23.41} & \textbf{0.846} & \textbf{0.305} \\
\bottomrule
\end{tabular}
\label{tab:kitti}
\vspace{-3ex}
\end{table}

\begin{figure}[t!]
\setlength{\abovecaptionskip}{1ex}
    \centering
    \includegraphics[width=0.90\linewidth, trim=0cm 0cm 0cm 0cm, clip]{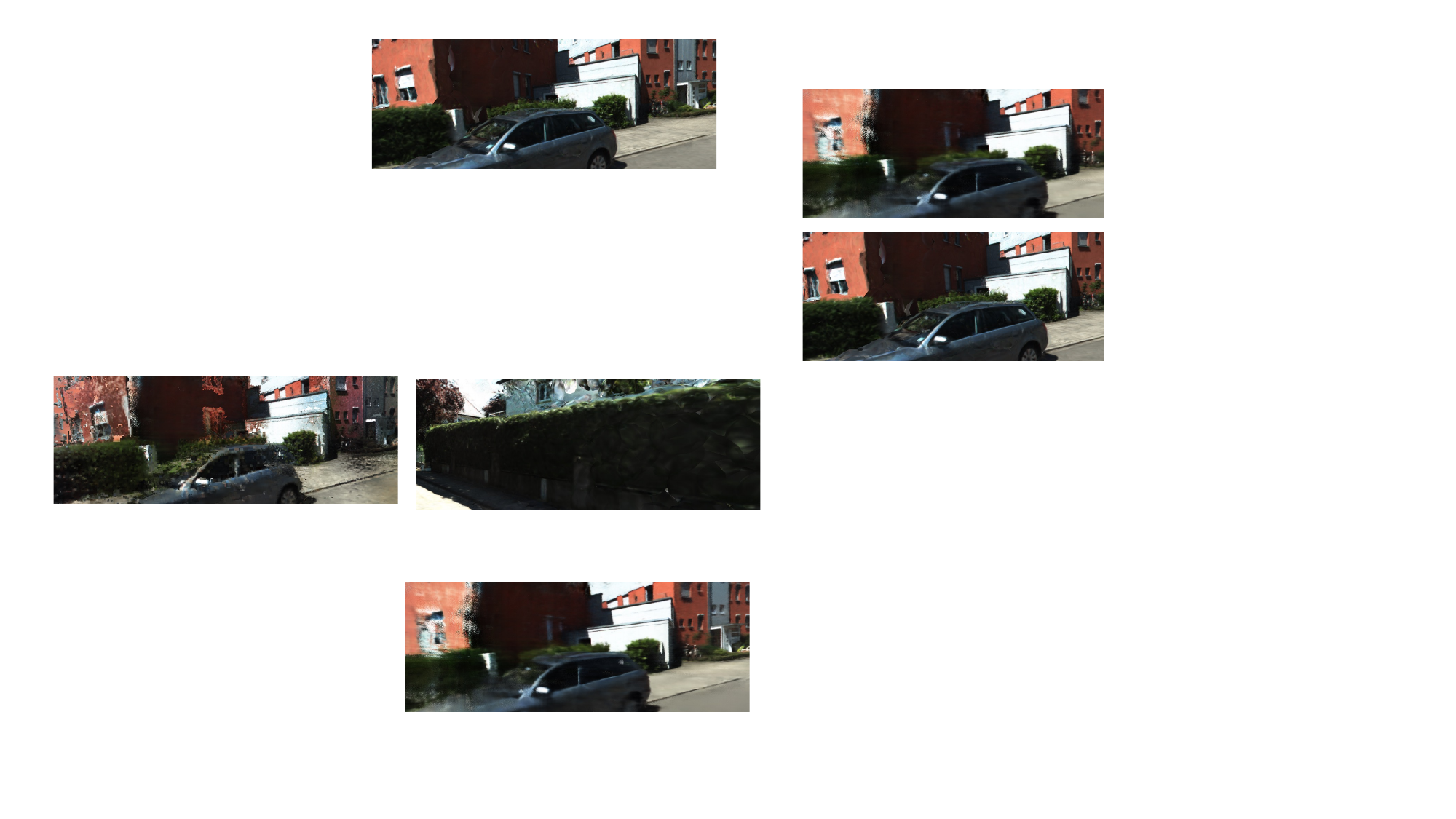}
    \caption{Qualitative comparison of novel view synthesis with significant view differences from the training set. Compared to Mip-NeRF and Point-NeRF, the proposed method can produce high-quality results with much less blurs and artifacts. Better zoom in for more details.}
    \label{fig:extrapolation}
    \vspace{-2ex}
\end{figure}

\begin{table}[t]
\centering
\setlength\tabcolsep{4pt} 
\renewcommand\arraystretch{1.15}
\caption{
Performance comparison of novel view synthesis with other competitive baselines on  Waymo dataset.
}
\footnotesize
\begin{tabular}{l c c c}
\toprule
\textbf{Method} & PSNR$\uparrow$ & SSIM$\uparrow$ & LPIPS$\downarrow$ \\
\midrule
NeRF~\cite{mildenhall2021nerf} & 26.24 & 0.870 & 0.472 \\
NeRF-W~\cite{martin2021nerf} & 26.92 & 0.885 & 0.418 \\
Instant-NGP~\cite{mueller2022instant} & 26.77 & 0.887 & 0.401 \\
Point-NeRF~\cite{xu2022point} & 26.26 & 0.868 & 0.450 \\
NPLF~\cite{ost2022neural} & 25.62 & 0.879 & 0.450 \\
Mip-NeRF~\cite{barron2021mip} & 26.96 & 0.880 & 0.451 \\
Mip-NeRF 360~\cite{barron2022mip} & \underline{27.43} & \textbf{0.893} & \underline{0.394} \\
\midrule
Ours & \textbf{27.62} & \underline{0.892} & \textbf{0.381} \\      
\bottomrule
\end{tabular}
\label{tab:waymo}
\vspace{-3ex}
\end{table}
To further evaluate the view synthesis ability for the views that are significantly different from the training set,  
we rotate the original testing views of KITTI-360 dataset to the left or the right by $45^\circ$ for evaluation. 
The performance comparisons under this setting are shown in  Table~\ref{tab:extrapolation}.  
Due to the explicit surface modeling with DNMPs, our method demonstrates relatively stronger robustness against the viewpoint changes compared with Mip-NeRF and Point-NeRF.     
According to the qualitative results in Fig.~\ref{fig:extrapolation}, our method can still output reasonable rendering results under $45^\circ$ view rotations, while the counterparts have failed and generated severe blur and artifacts.       

\begin{table}[t]
\centering

\caption{
View synthesis quality for the views that are significantly different from the training set. 
PSNR is adopted as the evaluation metric and the experiments are based on KITTI-360 dataset. 
}
\label{tab:extrapolation}

\footnotesize
\setlength{\tabcolsep}{3.2mm}{
{
\begin{tabular}{l c c c}
\toprule
\multirow{1}{*}{\textbf{Method}} & \multicolumn{1}{c}{Point-NeRF~\cite{xu2022point}} & \multicolumn{1}{c}{Mip-NeRF~\cite{barron2021mip}} & \multicolumn{1}{c}{Ours}\\
\midrule
$45^{\circ}$ left rot & 12.63 & 14.33 & \textbf{15.25} \\  
$45^{\circ}$ right rot & 14.32 & 16.16 & \textbf{17.09} \\                 
\bottomrule
\end{tabular}
}
}
\vspace{-3ex}
\end{table}

\subsection{Efficiency Analysis} 
\begin{table}[t]
\centering
\setlength\tabcolsep{4pt} 
\renewcommand\arraystretch{1.15}
\caption{Efficiency analysis. }

\resizebox{0.9\linewidth}{!}{
\begin{tabular}{l c c c c c c c c c c c c c c c c c c c }
\toprule
\textbf{Methods} & \makecell[c]{\# Network evaluations\\(/pixel)} & \makecell[c]{Peak memory\\(GB/1K pixel)} & \makecell[c]{Rendering time\\(ms/1K pixel)} \\
\midrule
NeRF~\cite{mildenhall2021nerf} & 256 & 0.80 & 20.24 \\
Mip-NeRF 360~\cite{barron2022mip} & 96 & 0.53 & 10.39 \\
Point-NeRF~\cite{xu2022point} & 40 & 2.06 & 32.44 \\
NPLF~\cite{ost2022neural} & \textbf{1} & 0.10 & 20.08 \\
Instant-NGP~\cite{mueller2022instant} & $\sim$126 & \textbf{0.02} & 0.71 \\
Ours & 6 & 0.11 & 2.07 \\
Ours (Lightweight) & 6 & 0.03 & \textbf{0.61} \\
\bottomrule
\end{tabular}
}
\label{tab:runtime}
\vspace{-4ex}
\end{table}
We analyze the efficiency of our system in terms of rendering time and memory consumption, and compare them with other typical methods. 
All the evaluations are based
on an NVIDIA A100 GPU. The results are exhibited in   Table~\ref{tab:runtime}. 
Given our explicit surface modeling, we only need the 6 nearest surface points intersected by the view ray (4 in the finest hierarchy and 2 in the coarser one) and evaluate the network on them, which significantly reduces the runtime and GPU memory usage.
Besides, we leverage rasterization for radiance feature interpolation rather than the inefficient k-NN algorithm adopted by Point-NeRF~\cite{xu2022point} and NPLF~\cite{ost2022neural}, which further improves the efficiency.    
Consequently, we achieve $5\times$ faster rendering speed with only ${\sim}\frac{1}{5}$ peak memory consumption compared with Mip-NeRF 360.  
Instant-NGP~\cite{mueller2022instant} is well-known for its fast inference speed 
based on the highly-optimized implementation. 
Although our system is naively implemented with PyTorch,
we achieve an inference speed competitive with Instant-NGP thanks to the efficient rasterization-based rendering and the small network evaluation times. Besides, our lightweight version can achieve even faster speed while maintaining better rendering quality than Instant-NGP (as shown in Table~\ref{tab:ablation},~\ref{tab:kitti}).
\begin{figure}
\centering
{
 \includegraphics[width=0.95\linewidth]{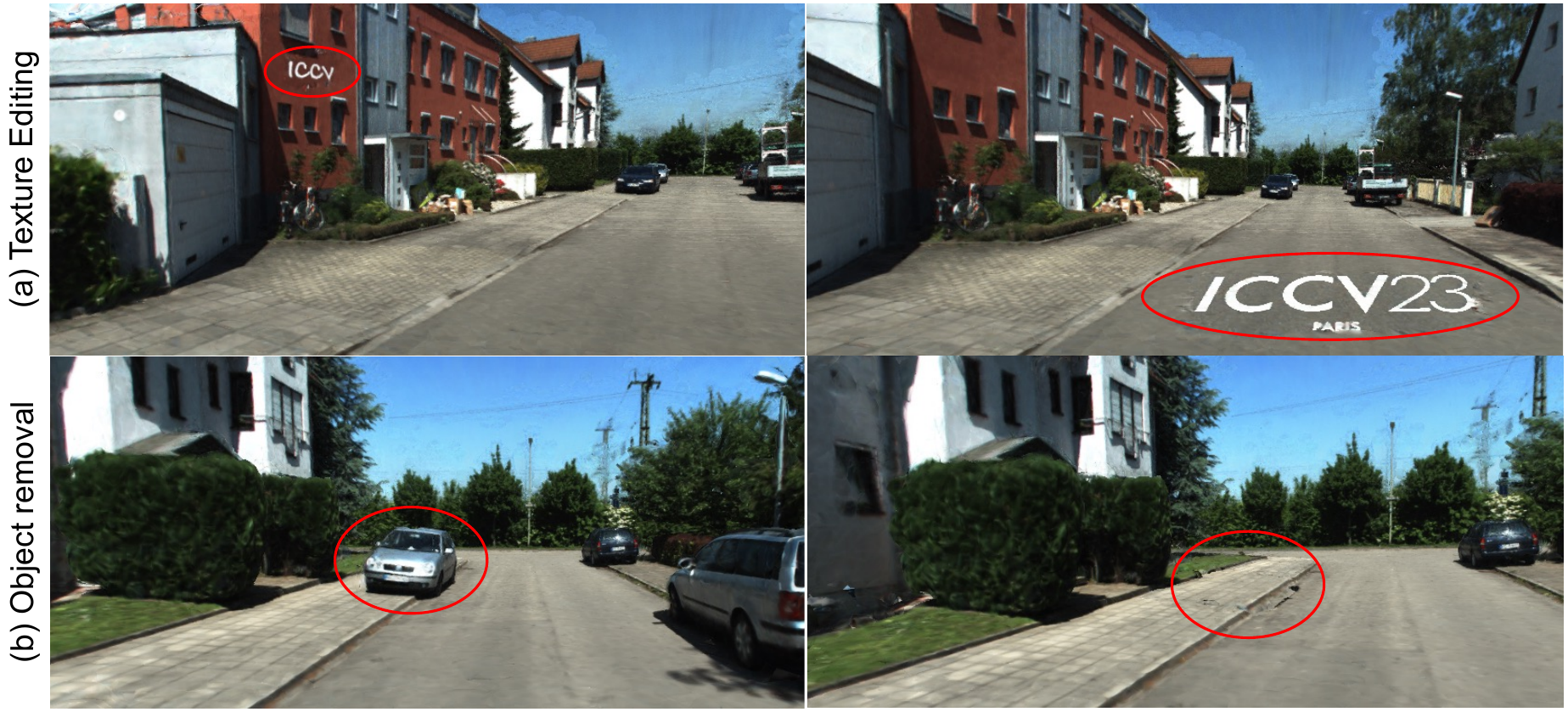}
 }
\caption{Samples of scene editing. Please refer to our supplementary materials for more visualization results.
}
\label{fig:scene_editing}
\vspace{-2ex}
\end{figure}
\subsection{Scene Editing} Our explicit mesh representation naturally enables scene editing. We can achieve texture editing (Fig.~\ref{fig:scene_editing} (a)) and object removal/insertion (Fig.~\ref{fig:scene_editing} (b)) by locally modifying the radiance features of vertices and removing/inserting the DNMPs. Please refer to our supplementary materials for more visualization results.

\section{Conclusion}
We have presented a novel neural scene representation for urban view synthesis based on the proposed Deformable Neural Mesh Primitives (DNMPs), which combines the efficiency of classic meshes and the representation capability of neural features. 
The entire scene is voxelized and each voxel is assigned a DNMP to  parameterize the geometry and radiance of the local area.     
The shape of DNMP is decoded from a compact latent space to constrain the degree of freedom for robust shape optimization. The radiance features are associated to each mesh vertex of DNMPs for radiance information encoding.       
Extensive experiments on two public outdoor datasets have verified the effectiveness of the proposed components and   
demonstrated the state-of-the-art performance of the proposed method. Moreover, due to the compact and efficient mesh-based representation, we achieve a 
fast inference speed and much lower peak memory compared with previous methods. 

However, although the remarkable rendering quality and resource efficiency have been achieved, the current version of our framework is still based on the static-scene assumption. In the future, we plan to extend our method to handle moving objects for more general application scenarios.     

{\small
\bibliographystyle{ieee_fullname}
\bibliography{egbib}
}

\end{document}